\documentclass[conference]{IEEEtran}
\IEEEoverridecommandlockouts
\usepackage{cite}
\usepackage{amsmath,amssymb,amsfonts}
\usepackage{algorithmic}
\usepackage{graphicx}
\usepackage{textcomp}
\usepackage{xcolor}
\usepackage{multirow}
\usepackage{array}
\usepackage{multicol}
\usepackage{etoolbox}
\usepackage{bm}
\usepackage{subfig}
\usepackage{etoolbox}
\usepackage{url}
\newenvironment{my_enumerate}{
\begin{itemize}
  \setlength{\itemsep}{1pt}
  \setlength{\parskip}{0pt}
  \setlength{\parsep}{0pt}}{\end{itemize}
}
\newcommand*{\thead}[1]{\multicolumn{1}{|c|}{\bfseries #1}}

\usepackage{amssymb}
\usepackage{pifont}
\newcommand{\xmark}{\ding{53}}%

\usepackage{tikz}
\def\ch{\tikz\fill[scale=0.4](0,.35) -- (.25,0) -- (1,.7) -- (.25,.15) -- cycle;}

\DeclareMathOperator*{\argmax}{arg\,max}  
\makeatletter
\patchcmd{\@makecaption}
  {\scshape}
  {}
  {}
  {}
\makeatother

\def\BibTeX{{\rm B\kern-.05em{\sc i\kern-.025em b}\kern-.08em
    T\kern-.1667em\lower.7ex\hbox{E}\kern-.125emX}}
\begin{document}

\title{Improving Neural Sequence Labelling using Additional Linguistic Information\\
{\footnotesize \textsuperscript{}}
}

\author{\IEEEauthorblockN{Mahtab Ahmed, Muhammad Rifayat Samee, Robert E. Mercer}
\IEEEauthorblockA{\textit{Department of Computer Science} \\
\textit{University of Western Ontario}\\
London, Ontario, Canada \\
mahme255, msamee, rmercer@uwo.ca}}

\maketitle

\begin{abstract}
Sequence labelling is the task of assigning categorical labels to a data sequence. In Natural Language Processing, sequence labelling can be applied to various fundamental problems, such as Part of Speech (POS) tagging, Named Entity Recognition (NER), and Chunking. In this study, we propose a method to add various linguistic features to the neural sequence framework to improve sequence labelling. Besides word level knowledge, sense embeddings are added to provide semantic information. Additionally, selective readings of character embeddings are added to capture contextual as well as morphological features for each word in a sentence. Compared to previous methods, these added linguistic features allow us to design a more concise model and perform more efficient training. Our proposed architecture achieves state of the art results on the benchmark datasets of POS, NER, and chunking. Moreover, the convergence rate of our model is significantly better than the previous state of the art models.
\end{abstract}


\section{Introduction}

Linguistic sequence labelling is one of the first tasks focusing on natural language processing using deep learning and it has been well examined over the past decade \cite{collobert2008unified, li2015tree, zheng2013deep}. Part of speech (POS) tagging, named entity recognition (NER), and chunking are subclasses of sequence labelling. They play a vital role in fulfilling many downstream applications,
such as relation extraction, syntactic parsing, and entity linking \cite{ratinov2009design, ando2005high, yarowsky1994decision}. POS tagging assigns a tag to each word in a text, where a tag represents the lexical category of a word. NER is a subtask of information extraction that seeks to locate and classify named entities in text. Chunking identifies the POS and short phrases in a sentence by doing shallow parsing and also groups words into syntactically correlated phrases. These labelled texts can later be used for different applications such as machine translation, information retrieval, word sense disambiguation, and natural language understanding etc.  

Before neural sequence models, traditional algorithms were based on Hidden Markov Models (HMMs) \cite{ekbal2007pos,zhou2002named} and Conditional Random Fields (CRFs) \cite{lafferty2001conditional, nguyen2007comparisons}. The problem with these models is they are heavily dependent on manually hand-crafted features. So it becomes difficult to apply them in real life applications because it is not practically possible to always have human expertise. 

To overcome these drawbacks, Neural Network (NN) based models have been proposed in which the models are responsible for extracting higher level features from the data \cite{benediktsson1990neural, kalchbrenner2014convolutional}. Recurrent Neural Networks (RNNs) along with its variants, Long Short Term Memory (LSTM) \cite{hochreiter1997long, cho2014learning} and Gated Recurrent Units (GRUs) are found to work very well with sequence data as they can capture long distance dependencies \cite{graves2006connectionist, cho2014learning, sutskever2014sequence}. Nevertheless, considering the overwhelming number of their parameters and the relatively small size of most human annotated sequence labelling corpora, annotations alone may not be sufficient to train complicated models. So, guiding the learning process with extra knowledge could be a wise choice \cite{manning2011part, sennrich2016linguistic}. For example, before tagging the word `\textit{flies}' as either a verb or a noun in the sentence `\textit{Time flies like an arrow}', having its semantic meaning would make a correct tagging straightforward. 

Knowing the sense of a word prior to sequence labelling (POS or NER) often gives the most probable tag for that word. Word senses can be obtained from a variety of sources: WordNet \cite{miller1995wordnet}, a lexical database for English that can be queried for the sense of a word given its context; the simplified  LESK algorithm \cite{basile2014enhanced, banerjee2002adapted} which uses the dictionary  definition of each word in a sentence as extra context to suggest the word sense; and linear algebraic methods, one \cite{arora2016linear} which uses a random walk on a discourse model and represents the vector of the base word as a linear combination of its probable sense vectors. 

In this paper, we propose a novel deep neural architecture for doing sequence labelling incorporating not only semantic features through word senses but also the rich morphology of the words. We provide an in depth analysis of the design of this architecture giving some insights regarding how each feature is introduced into the architecture. Our sequence model achieves state of the art results on the three sequence labelling tasks, POS, NER, and chunking, and has a training time at least four times faster than the currently available state of the art models. 

\section{Related work}
Huang et al.\ \cite{huang2015bidirectional} propose a few models for the sequence tagging task. Apart from just word embeddings, they first use morphological as well as bigram and trigram information as their input features. Later, they use LSTM and Bidirectional LSTM (BLSTM) with CRF to do the final tagging. Lample et al.\ \cite{lample2016neural} extract character embeddings from both the left and right directions, concatenate these with word embeddings and then use a stacked LSTM along with CRF to do the tagging. Liu et al.\ \cite{liu2017empower} propose a model leveraging both word as well as character level features. It includes a language model to represent the character level knowledge along with a highway layer to avoid the feature collision. Finally it is trained jointly as a multitask learning.

Yu et al.\ \cite{yu2017general} propose a general purpose tagger using a convolutional neural network (CNN). First, they use CNNs to extract the character level features and then concatenate it with word embeddings, position embeddings and binary features. Finally they use another CNN to get the contextual features as well as to do the tagging. Ma and Hovy \cite{ma2016end} propose an end-to-end sequence labelling model using a combination of BLSTM, CNN and CRF. They use a CNN to get the character level information, concatenate it with word embeddings and then apply BLSTM to model the contextual information. Finally they generate the tags by using a sequential CRF layer. Rei \cite{rei2017semi} trains a language model type objective function using BLSTM-CRF to predict the surrounding words for every word in the corpus and utilizes it for sequence labelling.

The contribution of this paper combines the common themes found in these previous works (morphology encoded as character embeddings, and word embeddings) with word senses in a new architecture that integrates these embeddings and the outputs of a CNN, a BLSTM, and a CRF in novel ways.

\section{The Model: BLSTM-CRF}

In this section, we describe our work in detail. We first explain each of the pieces of the complete architecture and then we explain how we combine those pieces to build our model. This section also explains the morphological and semantic features that we have added with our model to get the improved performance that is discussed in Section \ref{results}.

\subsection{Recurrent unit: Bidirectional LSTM}
Recurrent neural networks (RNNs) are the best known and most widely used NN model for sequence data as they go over the entire sequence through time and try to remember it in a compressed form. 
Although its variant, LSTM, is very good with long term dependencies, for many sequence labelling task, it is important to keep track of these dependencies from the future as well as from the past. But LSTM has just one hidden state from the past and changes that hidden state recursively through time. An elegant solution to this problem is going over the sequence in both forward and backward directions with two hidden states and finally concatenating the output from both directions. This bidirectionality has proven to be very effective in some prior works \cite{graves2005framewise, thireou2007bidirectional,  dyer2015transition}. The resulting network, the Bidirectional LSTM (BLSTM), is the RNN variant that is used by the model described below.

\subsection{Word Sense}

Knowing the sense of a word prior to tagging makes the tagging task more straightforward. Generally, polysemy is captured in standard word vectors, but the senses are not represented as multiple vectors. So we have trained an adaptive skip gram model, AdaGram, \cite{bartunov2016breaking} which gives a vector for each sense of a word. It is a non-parametric Bayesian extension of the skip-gram model and is based on the constructive definition of Dirichlet process (DP) \cite{ferguson1973bayesian}. It can learn the required number of representations of a word automatically. 

In our model, we denote a set of input words as $X= \{x_{i}\}_{i=1}^{N}$ and their context as $Y= \{y_{i}\}_{i=1}^{N}$. The \textit{i}th training pair $(x_i, y_i)$ consists of words $x_i = o_i$ with context $y_i = (o_t)_{t \in c(i)}$, where $c(i)$ is the index of the context words. Then, instead of maximizing the probability of generating a word given its contexts \cite{mikolov2013distributed}, we maximize the probability of generating the context given its corresponding input words \cite{bartunov2016breaking}. Our final objective function becomes,

\begin{equation} \label{first_equ}
    p(Y|X,\theta) = \prod_{i=1}^{N} p(y_i|x_i, \theta) = \prod_{i=1}^{N}\prod_{j=1}^{C} p(y_{ij}|x_i, \theta)
\end{equation}
where, $\theta$ is the set of model parameters. The drawbacks of this objective function is that it captures just one representation of a word which goes against a word having different senses depending on the context \cite{arora2016linear}. To counter this, AdaGram introduces a new latent variable $z$  which captures the required number of senses even though the number of structure components of the data is unknown a priori. In AdaGram, if the similarities of a word vector with all its existing sense vectors are below a certain threshold, a new sense is assigned to that word with a prior probability $p$. The prior probability of the $k$th meaning of word $w$ is 
\begin{equation} \label{second_equ}
\begin{split}
p(z=k|w,\beta)=\beta_{wk}\prod_{r=1}^{k-1}(1-\beta_{wr}), \\ p(\beta_{wk}|\alpha)=Beta(\beta_{wk}1,\alpha), 
k=1\ldots 
\end{split}
\end{equation}
where $\beta$ is a latent variable and $\alpha$ controls the number of senses. Theoretically, it is possible to have an infinite number of senses for each word $w$. However, as long as we have a finite amount of data, the number of senses can not be more than the number of occurrences of that word. With more data, it can increase the complexity of the latent variables thereby allowing more distinctive meanings to be captured. Taking all the facts into account, our final objective function becomes, 
\begin{equation} \label{third_equ}
\begin{split}
p(Y,Z,\beta|X,\alpha, \theta)=\prod_{w=1}^{V}\prod_{k=1}^{\infty}p(\beta_{wk}|\alpha) \\ \prod_{i=1}^{N}[p(z_{i}|x_i,\beta)\prod_{j=1}^{C}[p(y_{ij}|z_i,x_i,\theta)]
\end{split}
\end{equation}
where $Z = \{z_i\}_{i=1}^{N}$ is a set of senses for all the words.
\subsection{Convolutional Neural Network} \label{bigram}

Convolutional Neural Networks (CNNs) are good for extracting n-gram features from a sentence \cite{lei2015molding}. They consist of kernels (i.e., a matrix of weights) which are used to go over the input word embedding matrix with a variable stride length and extract some higher level features. In our model, we use a CNN to get the bigram features. To do that, first we pad the input sentence \footnote{The `$valid$' convolution operation reduces the size of the feature matrix by $k-L$, where $k$ is the size of the kernel vector and $L$ is the stride length. For us, $L = 1$. To keep the size of the feature matrix uniform through the model, we padded a start token $\textless start \textgreater$ at the beginning of the input sentence.} and pass it to an embedding layer. This layer represents the sentence as a matrix of size $(m+1) \times d$, where $m$ is the actual sentence length and $d$ is the embedding dimension. Next, we initialize a kernel of size $2 \times d$ with stride length $1$ and convolve it with the input sentence matrix. This results in a bigram embedding matrix $B$ of size $m \times d$ using Eqn. \ref{fourth-equ}.
\begin{equation}  \label{fourth-equ}
B_{i,:}=\ \sum^2_{j=1}{I_{i+j,:}*K_{j,:}} 
\end{equation}  
where $I$ is the input sentence matrix, $K$ is the convolution kernel and $n$ is the maximum sequence length for the current batch. Later, this bigram embedding is passed to a BLSTM layer to extract  more abstract features.

\subsection{Conditional Random Field}
Each of the tasks that we are modelling requires a tag to be assigned to each word. In addition to using the current word to predict its tag, it is also possible to use the information about the neighboring words' tags. There are two main ways to do this. One way is to calculate the distribution of tags over each time step and then use a beam search-like algorithm, such as maximum entropy markov models \cite{mccallum2000maximum} and maximum entropy classifiers \cite{ratnaparkhi1996maximum}, to find the optimal sequence. Another way is to focus on the entire sentence rather than just the specific positions which leads to Conditional Random Fields (CRFs) \cite{lafferty2001conditional}. CRFs have proven to give a higher tagging accuracy in cases where there are dependencies between the labels. Like the bidirectionality of BLSTM networks a CRF can provide tagging information by looking at its input features bidirectionally.

In our model we denote a generic input sequence as $x= \{x_{i}\}_{i=1}^{N}$, generic tag sequence as $y= \{y_{i}\}_{i=1}^{N}$, and set of possible tag sequences of $x$ as $F(x)$. Then we use CRF to calculate the conditional probability over all possible tag sequences $y$ given $x$ as 
\begin{equation}  \label{fifth-equ}
p(y|x;W,b) = \frac{\prod_{i=1}^{n}\phi_i(y_{i-1},y_i;x)}{\sum_{y'\in F(x)}^{}{\prod_{i=1}^{n}\phi_i(y'_{i-1},y'_i;x)}}
\end{equation}  
where $\phi(.)$ is the score function for the transition between the tag pair $(y',y)$ given $x$. We train this CRF model using maximum likelihood estimation (MLE) \cite{johansen1990maximum}. For a training pair $(x_i, y_i)$ we maximize 
\begin{equation}  \label{sixth-equ}
\textit{L}(W,b) = \sum_{i}^{}\log p(y|x;W,b)
\end{equation}  
where $W$ is the weight matrix and $b$ is the bias term. While decoding, we search for the best tag $\hat{y}$ with the highest conditional probability using the Viterbi algorithm \cite{sha2003shallow}.
\begin{equation}  \label{seventh-equ}
\hat{y} = \argmax_{y \in F(x)}p(y|x;W,b)
\end{equation}  

\subsection{Morphology: Spelling and suffix features} \label{spellings}\label{suffix}
For the morphological features, we have focused on spelling features and suffix features.

We extract the following 14 spelling features for a given word and store it as a binary vector $SV_{1 \times 14}$.

\begin{my_enumerate}
  \item whether it is composed only of alphabetics or not
  \item whether it contains non-alphabetic characters except `.' or not
  \item whether it starts with a capital letter or not
  \item whether it is composed only of upper case letters or not
  \item whether it is composed only of lower case letters or not
  \item whether it is composed only of digits or not
  \item whether it is composed of a mixture of alphabetics and numbers or not
  \item whether it is the starting word in the sentence or not
  \item whether it is the last word in the sentence or not
  \item whether it is in the middle of the sentence or not
  \item whether it ends with an apostrophe s ('s) or not
  \item whether it has punctuation or not
  \item whether it is the first word in the sentence and starts with a capital letter or not 
  \item whether it is composed mostly of digits or not
\end{my_enumerate}
Apart from extracting these features, we also replace all the numbers in the corpus with the \textless \textit{number}\textgreater tag. 

We have assembled a list of $137$ suffixes from \url{https://www.learnthat.org/pages/view/suffix.html} and have used the ten that occur most often in our corpus for this study. Then for each of these suffixes, we have collected the words that end with that suffix and have recorded their POSs as well as the frequency. Next, we made an assumption that if a word $w$ with POS $x$ ends with a specific suffix $s$ exceeds a frequency threshold in the training set, then $s$ is the true suffix of word $w$. We record the pair as $(w,s)$. Finally, we create a one hot vector $SUV_{1 \times 10}$ for each word where a $1$ at index $k$ means the word has the $k$th suffix.

\subsection{BLSTM-CRF model}
In this sub-section, we combine the BLSTM and CRF models with some feature connection techniques to form our final BLSTM-CRF model. We divided this final model into some modules and the description of each of these modules is as follows:

\textbf{Module 1: Word Level Features} This module starts with an embedding layer. In detail, we initialize the emebedding layer randomly as well as using pre-trained embeddings (GloVe / word2vec). Next we represent each sentence as a column vector $I_{m \times 1}$ where each element of the vector is a unique index of the corresponding word. Then we pass this vector to an embedding layer which gives a matrix representation $W_{m \times d}$. Here, $d$ is the embedding dimension. 

\textbf{Module 2: Character Embedding} In this module, first we split a word into its characters and then transform it into a column vector $C_{k \times 1}$, where $k$ is the word length and each element of the vector is a unique index of the corresponding character. Next we initialize an embedding layer randomly and pass the character vector into it. This will change the representation to a matrix of size $k \times n$ where  $n$ represents the embedding dimension. Then we use an LSTM on this matrix and store the last hidden state of this LSTM as the character level representation $C_{1 \times n}$ of the word. Finally, for a sentence with $m$ words, it is stored as a matrix $C_{m \times n}$.

\textbf{Module 3: Selective Pickup from Char LSTM (SP-CLSTM)} In this module, we introduce a new way of capturing the morphological features as well as the context features. The word embeddings from module $1$ gives the contextual features in both directions and the character embeddings from module $2$ gives the lexical information. We capture both sets of information by first representing each sentence in terms of its characters $I_{(k\times m) \times 1}$ and then turn this into a matrix of size $k \times m \times d$ through a random embedding layer. Then we apply a BLSTM over this representation and finally we pick those indices from the output where each word ends. This selective pickup provides the morphological information of a word as well as information about the previous words in the sequence.
\begin{equation}
    \tilde{C}_{m \times d} = \texttt{SELECT} (\texttt{BLSTM}(I_{(k\times m) \times d}))
\end{equation}

\textbf{Module 4: Sense Features} This module calculates the sense level contextual features of a sentence. First, we initialize a sense embedding layer using the pre-trained sense embeddings from AdaGram. Then we tag each word in the input sentence using the module \texttt{disambiguate} from AdaGram (the word `\textit{apple}' with sense $2$ is tagged as `\textit{apple}\textunderscore{2}'). This modified input sentence is then passed to the embedding layer initialized before and finally the resultant output is passed to a BLSTM layer. The output of this BLSTM layer gives the sense level contextual feature $S_{m \times d}$.

\textbf{Module 5: Bigram Features} This module calculates the bigram embedding features $B_{m \times d}$ of a sentence as described in the Subsection \ref{bigram}.

\begin{figure*}[hbtp]
\centering
  \includegraphics[width=15.8cm,height=22.8cm,angle=180]{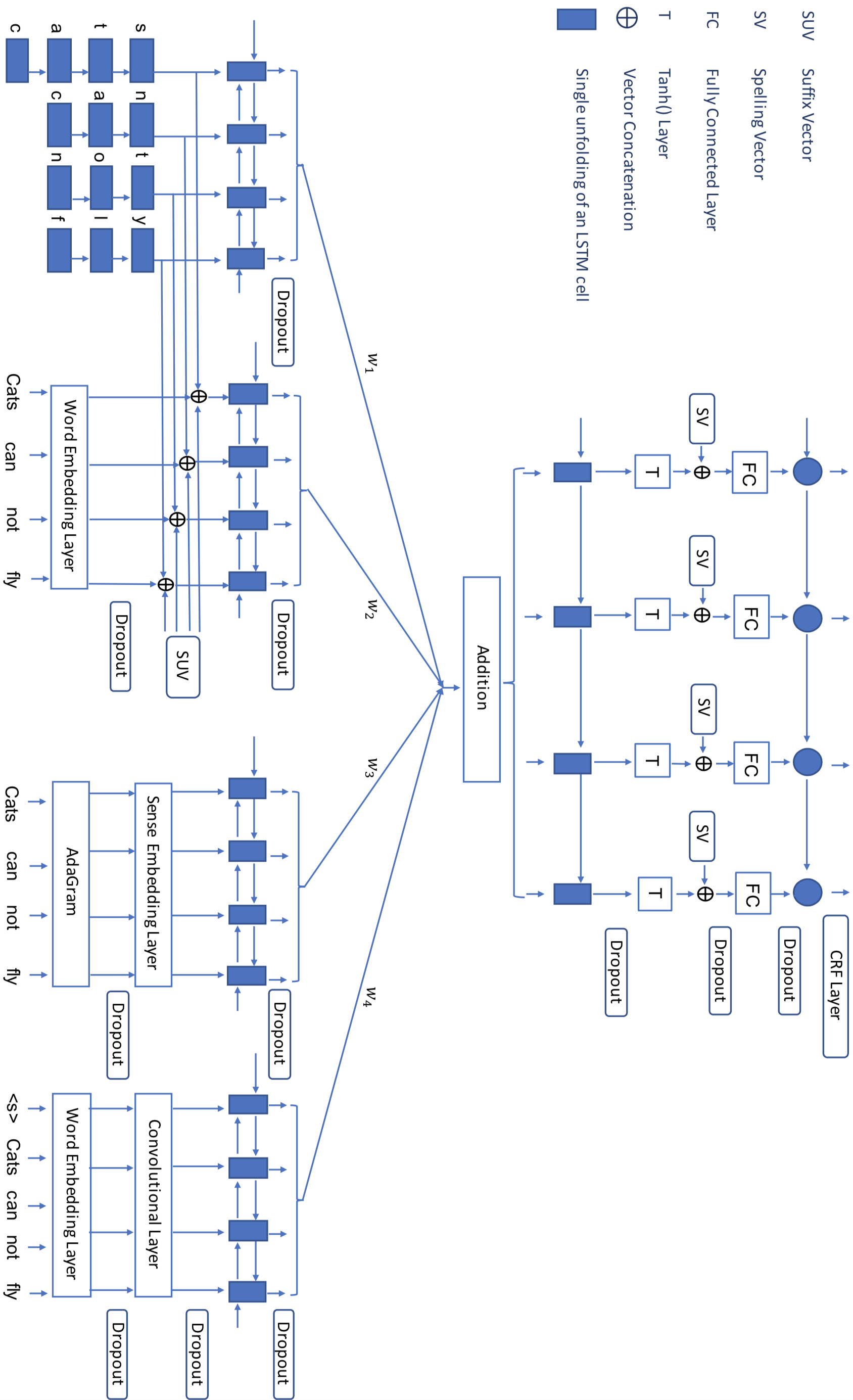}
  \caption{\label{model} BLSTM-CRF model architecture}
\end{figure*}

\textbf{Module 6: The Connection Technique} In this module, we combine all the features and the modules using some novel connection techniques and build our final BLSTM-CRF model as shown in Figure \ref{model}. First we concatenate the word embedding from module 1 with the character embedding from module 2 and the suffix vector from Subsection \ref{suffix} as $[W_{m \times d}, C_{m \times n}, SUV_{m \times 10} ]$. Following this, we apply a BLSTM on this new embedding matrix, calling this output $O^{1}_{m \times d}$. The outputs of modules 3, 4 and 5 are called $O^{2}_{m \times d}$, $O^{3}_{m \times d}$, and $O^{4}_{m \times d}$, respectively. Then we initialize four scalar weights $w_1$, $w_2$, $w_3$ and $w_4$ with initial value $1.0$ and add them as model parameters. We form a linear combination of the $w_i$ weighted $O^{i}$s to form the final output.
\begin{equation} \label{w1234}
    O = \sum_{i=1}^{4} O^{i} w_i
\end{equation}

The final output ($O_{m \times d}$) have pieces of information from all the features that we calculated above. We choose linear addition rather than concatenation of these output features, because concatenation will result in a very large feature matrix and the network have to tune each of the cell of this matrix during back-propagation. Following this, we initialize an LSTM layer where we pass the final output from Eqn. \ref{w1234} at each time step $\tilde{O}^{i}_{1 \times z} = \texttt{LSTM}(O^{i}_{1 \times d}, h^{i-1}_{1 \times d})$ and store the outputs separately  $\tilde{O}_{m \times z} = [\tilde{O}^{1}, \tilde{O}^{2}, \dots , \tilde{O}^{m}]$. This LSTM layer unfolds at each time step taking the hidden state of the previous time step to initialize the hidden state of the current time step. The previous hidden state has the information about the previous tag and initializing the current hidden state with the previous one explicitly gives this information. Next we pass the output from each time step to a \texttt{tanh} layer $T_{1 \times d} = \texttt{tanh}(\tilde{O}^{i})$, which squeezes the values between $[-1,1]$. Then we concatenate this \texttt{tanh} output $T_{1 \times d}$ with the spelling features $SF_{1 \times 14}$ calculated in subsection \ref{spellings} and pass this to a fully connected (FC) layer. This FC layer maps the output to the number of tag classes $Y_{1 \times c} = \texttt{FC} ([T_{1 \times d}, SUV_{1 \times 14}])$, where $c$ represents number of classes. We do this for each time step and concatenate the results to make a final tensor $Y_{m \times c}$. Finally, we pass this tensor to the CRF layer and calculate the possible tag sequence for the given input sequence. 

\section{Experimental Setup}
In  this  section,  we describe  the  detailed  experimental setup for the evaluation of our study. We first explain the dataset statistics for each tagging task. Following this, we explain the working environment details along with the hyper-parameter settings of our architecture.

\subsection{Dataset Description}

We test our BLSTM-CRF model on three NLP tagging tasks: Penn TreeBank (PTB) POS tagging, CoNLL 2000 chunking, and CoNLL 2003 NER. Table \ref{dataset} shows the number of sentences in the training, validation and test sets respectively for each corpus. We utilize the BIO2 explanation standard for the chunking and NER tasks.

\begin{table}[h]
\small
\centering
\begin{tabular}{ c | c | c | c } \hline 
 & \textbf{WSJ} & \textbf{CoNLL00} & \textbf{CoNLL03} \\ \hline
 
 Train & 39831 & 8936 & 14987 \\ \hline
 Valid & 1699 & N/A & 3466 \\ \hline
 Test & 2415 & 2012 & 3684 \\ \hline
\end{tabular}
\caption{\label{dataset}  Dataset Description }
\end{table}

\begin{table}[h]
\centering
\small
\begin{tabular}{ p{1.5in} | p{1.2 in} } \hline 
\multicolumn{2}{c}{\textbf{BLSTM-CRF}}\\ \hline 
\textbf{Hyper-parameter} & \textbf{Range Selected} \\ \hline 
Learning rate & 0.001 / 0.015 / 0.01 \\ \hline 
Batch size & 10 / 50 / 100 \\ \hline 
No. of LSTM layers  & 1 / 2 / 3 \\ \hline 
Momentum & 0.9 \\ \hline 
Dropout & 0.5 / 0.2 / 0.1 \\ \hline 
Word embedding size & 300 / 200 / 100 \\ \hline 
Character embedding size & 50 / 30  \\ \hline 
Initial scalar weight value & 1.0 \\ \hline 
Gradient clipping & 5 / 20 / 50 \\ \hline 
Weight decay & $10^{-5}$\\ \hline
Learning rate decay & 0.05\\ \hline
CNN kernel size & $2 \times (300/200/100)$ \\ \hline
\multicolumn{2}{c}{\textbf{AdaGram}} \\\hline 
Epoch & 1000 \\ \hline
Window size & 10 / 7 / 5 \\ \hline
No. of prototypes  & 5 \\ \hline 
Sense embedding size & 300 \\ \hline 
Prior prob. of new sense & 0.1 \\ \hline
Initial weight on first sense & -1 \\\hline
Word embedding size & 300 / 200 / 100 \\ \hline 
\end{tabular}
\caption{\label{hyper}  Ranges of different hyper-parameters searched during tuning. }
\end{table}
Table \ref{hyper} shows the detailed hyper-parameter settings of our model and some of the hyper-parameters for AdaGram (the remaining parameters are set to their default values \cite{bartunov2016git}). We train our model on Nvidia GeForce GTX 1080 GPU with both the `Adam' and `SGD' optimizers. All of the results in the next section are reported using `SGD' as it was giving the best results. The `Learning rate decay' parameter was only used with the `SGD' optimizer. We used PyTorch 0.3.1 to implement our model and Julia 0.4.5 for running AdaGram under the Linux environment.

\section{Experimental Results}
\label{results}

\begin{table}[b!]
\centering
\small
\begin{tabular}{ p{2.3in}| c } \hline 
\textbf{Model}  & \textbf{F1-score}\\ \hline
SVM classifier \cite{kudoh2000use} & 93.48 \\ \hline
SVM classifier \cite{kudo2001chunking} & 93.91 \\ \hline
BI-LSTM-CRF  \cite{huang2015bidirectional} & 94.13 \\ \hline
Second order CRF \cite{mcdonald2005flexible} & 94.29\\ \hline
Second order CRF \cite{sha2003shallow} & 94.30 \\ \hline
Conv. network tagger \cite{collobert2011natural} & 94.32\\ \hline
Second order CRF \cite{sun2008modeling} & 94.34\\ \hline
BLSTM-CRF (Senna) \cite{huang2015bidirectional} & 94.46 \\ \hline
HMM + voting \cite{shen2005voting} & 95.23\\ \hline \hline
BLSTM-CRF (Ours) & \textbf{96.76}
\end{tabular}
\caption{\label{chunk} Comparison of F1 scores of different models for chunking}
\end{table}
In this section, we describe in detail the results obtained with our proposed architecture. As the evaluation metrics, we use accuracy for the WSJ corpus and F1 score (micro averaged) for the CoNLL00 and CoNLL03 tasks. This section also contains the results of the top performing models for all three sequence labelling tasks. Additionally, we show the rate of convergence of our model compared to the state of the art one. Finally, we conclude this section by giving an ablation study by removing certain modules as well as features and mixing them in different combinations.
\begin{table}[t]
\centering
\small
\begin{tabular}{ p{2.3in}| c } \hline 
\textbf{Model}  & \textbf{F1-score}\\ \hline
Conv-CRF \cite{collobert2011natural} & 81.47 \\ \hline
BLSTM-CRF \cite{huang2015bidirectional} & 84.26 \\ \hline
MaxEnt classifier \cite{chieu2002named} &  88.31 \\ \hline
HMM + Maxent \cite{florian2003named} & 88.76 \\ \hline
Semi-supervised \cite{ando2005framework} & 89.31 \\ \hline
Conv-CRF + Senna \cite{collobert2011natural} & 89.59 \\ \hline
BLSTM-CRF \cite{huang2015bidirectional} & 90.10 \\ \hline
CRF + LIE \cite{passos2014lexicon} & 90.90 \\ \hline \hline
BLSTM-CRF (Ours) & \textbf{91.63} \\ \hline
\end{tabular}
\caption{\label{NER} Comparison of F1 scores of different models for NER}
\end{table}

Table \ref{chunk} shows the performances of all of the chunking systems. An SVM based classifier \cite{kudoh2000use} won the CoNLL 2000 challenge with an F1 score of 93.48\%. However, later they improved their result up to 93.91\% \cite{kudo2001chunking}. Recently, most of the models incorporate CRF in their architecture to capture the tag dependencies and achieve very good performance \cite{sha2003shallow, mcdonald2005flexible, sun2008modeling}. However, none of them surpass the performance of \cite{shen2005voting} which uses an HMM to capture the dependencies and a voting scheme to increase the confidence interval of the model. Our model outperforms all the existing models and achieves a state of the art F1 score of \textbf{96.76\%}.

\begin{table}[b!]
\centering
\small
\begin{tabular}{ p{2.3in}| c } \hline 
\textbf{Model}  & \textbf{Accuracy}\\ \hline
Conv-CRF \cite{collobert2011natural} &  97.29 \\ \hline
5wShapesDS \cite{manning2011part} & 97.32 \\ \hline
Structure regularization \cite{sun2014structure} & 97.36 \\ \hline
Multitask learning \cite{rei2017semi} & 97.43 \\ \hline
Nearest neighbor \cite{sogaard2011semisupervised} & 97.50 \\ \hline
LSTM-CRF \cite{lample2016neural} & 97.51 \\ \hline
LSTM-CNN-CRF \cite{ma2016end} &  97.55 \\ \hline
LM-LSTM-CRF \cite{liu2017empower} & 97.59 \\ \hline \hline
BLSTM-CRF (Ours) & 97.51 \\ \hline
BLSTM-CRF (Ours) without CNN & \textbf{97.58} \\ \hline
\end{tabular}
\caption{\label{POS} Comparison of Accuracy of different models for POS tagging}
\end{table}
\begin{table*}[!h]
\centering
\small
\begin{tabular}{ c|c|c|c|c|c|c|c|c|c|c|c|c|c} \hline 
\textbf{Model} &\textbf{Word} & \multirow{2}{*}{\textbf{Sense}} & \multirow{2}{*}{\textbf{SP-CLSTM}} & \multirow{2}{*}{\textbf{Bigram}} & \multicolumn{2}{|c|}{\textbf{Suffix}}& \multicolumn{3}{|c|}{\textbf{Spelling}} & \multicolumn{2}{|c|}{\textbf{Char Embed}} & \textbf{Prev.} & \multirow{2}{*}{\textbf{Acc.}}\\ \cline{6-12}
 \textbf{no.}& \textbf{emb} & & &  &CW &CO &R&CW&CO &CW&CO &\textbf{POS}& \\ \hline
1 & Rand. & \xmark &\xmark & \xmark  &-&-&-&-&-&-&-&\xmark&95.42\\ \hline
2 & Glove &\xmark &\xmark & \xmark  &-&-&-&-&-&-&-&\xmark&96.13\\ \hline
3 & Glove &\ch &\xmark & \xmark  &-&-&-&\ch&-&-&\ch&\xmark&97.08\\ \hline
4 & Glove &\ch &\ch & \xmark  &\ch&-&\ch&-&-&\ch&-&\xmark&97.15\\ \hline
5 & Glove &\ch &\ch & \xmark  &\ch&-&-&\ch&-&\ch&-&\xmark&97.22\\ \hline
6 & Glove &\ch &\xmark & \xmark  &-&\ch&-&-&\ch&\ch&-&\xmark&97.32\\ \hline
7 & Glove &\ch &\ch & \xmark  &-&- &-&-&\ch &\ch&- & \xmark&97.45\\ \hline
8 & Glove &\ch &\ch & \xmark  &\ch&- &-&-&\ch &\ch&- & \xmark&97.48\\ \hline
9 & Glove &\ch &\ch & \ch  &\ch&- &-&-&\ch &\ch&- & \xmark&97.50\\ \hline
10 & Glove &\ch &\ch & \xmark  &\ch&-&-&-&\ch&\ch&-&\ch&\textbf{97.58}\\ \hline

\end{tabular}
\caption{\label{tricks} Ablation study of our BLSTM-CRF model for POS tagging. (R - Residual connection, CW - Concatenate with word embedding, CO - Concatenate with second last output layer)}
\end{table*}
Table \ref{NER} shows the results of the existing models on the NER task. Huang et al.\ \cite{huang2015bidirectional} did many experiments using random and pre-trained embeddings on their model. For random embeddings, they achieved a very low score of 84.26\%. However, when they use pre-trained SENNA embeddings \cite{collobert2011natural} along with a gazetteer feature, their F1-score jumped up to 90.10\% surpassing the Conv-CRF model \cite{collobert2011natural} which uses window and sequence approach networks to do the tagging. Our model achieves a state of the art result of \textbf{91.63\%}.

\begin{table}[t]
\centering
\small
\begin{tabular}{| p{0.2in}| p{1in} | c | c |} \hline 

\multicolumn{2}{|c|}{\textbf{Model}}  & \textbf{Acc.} & \textbf{Time}\\ \hline

\multirow{5}{*}{\cite{liu2017empower}}
& LSTM-CRF & 97.35 & 37 \\ \cline{2-4} 
& LSTM-CNN-CRF & 97.42 & 21 \\ \cline{2-4} 
& LM-LSTM-CRF  & 97.53 &16 \\ \cline{2-4} 
& LSTM-CRF & 97.44 & 8 \\ \cline{2-4} 
& LSTM-CNN-CRF & 96.98 & 7 \\ \hline \hline
\multirow{2}{*}{Ours}
& BLSTM-CRF & 97.58 & 4\\ \cline{2-4}
& BLSTM-CRF without CNN & 97.51 & 3.5 \\ \hline
\end{tabular}
\caption{\label{time}  Training time (in hours) of our BLSTM-CRF model on the WSJ corpus compared with all the models of \cite{liu2017empower} using the same hardware configuration (GPU: Nvidia GTX 1080)}
\end{table}

\begin{table}[ht]
\small
\centering
\begin{tabular}{|l|c|c|c|c|c|} \hline

\multirow{2}{*}{\textbf{Module}} &  \thead{100th} & \thead{200th} & \thead{300th} & \thead{400th} & \thead{500th} \\
\multirow{2}{*}{} &  \thead{epoch} & \thead{epoch} & \thead{epoch} & \thead{epoch} & \thead{epoch} \\ \hline
Word emb & 0.91 & 0.84 & 0.80 & 0.77 &  0.78 \\ \hline 
Sense & 0.85 & 0.76 & 0.69 & 0.64 & 0.65 \\ \hline 
SP-CLSTM & 0.81 & 0.66 & 0.48 & 0.35 & 0.34 \\ \hline 
Bigram & 0.75 & 0.49 & 0.27 & 0.01 & 0.01 \\ \hline 
\end{tabular}
\caption{\label{weights}  Change in $w$'s for each module with epochs.}
\end{table}

Table \ref{POS} shows the performance of our architecture in comparison with some top performing ones for the POS tagging task. As can be seen, a number of models use Convolution or LSTM or BLSTM to get the contextual features and CRF to do the tagging. They achieve very good accuracies of 97.29\% \cite{collobert2011natural},  97.51\% \cite{lample2016neural} and 97.55\% \cite{ma2016end}. Some of the models use multitask learning, doing two or more tasks at the same time. They also achieve very good accuracies: 97.43\% \cite{rei2017semi} and \textbf{97.59\%} \cite{liu2017empower}. Our model achieves an accuracy of \textbf{97.58\%} which is higher than all of the existing models except LM-LSTM-CRF \cite{liu2017empower} which leverages a language model for the tagging tasks. LM-LSTM-CRF, however, has a mean accuracy of \textbf{97.53\%} (reported accuracy: $97.53\pm{0.03}$) which is lower than the our model's mean accuracy ($97.57\pm{0.01}$). Also, as shown in Table \ref{time}, our model's training time is one quarter that of LM-LSTM-CRF with on par performance.

Table \ref{tricks} gives the ablation study of our model where we show how we apply different combinations of features in different parts of our BLSTM-CRF architecture to get an optimal configuration. With so many features and parameters, these sequence models are very much prone to overfit. But with careful tuning as well as with proper feature connections, it is possible to leverage those features. We extract a set of morphological as well as semantic features from our dataset such as spelling, suffix and char-level features. We experiment on applying various combinations of these features in different segments of our model. Our extensive experimentation shows that optimal results are achieved when these features are added in the model through residual connection, concatenation with word embeddings and concatenation with the second last output layer. Focusing on which segment to connect each feature, our experiments found that the spelling feature works best when concatenated with the second last output layer, and the suffix feature as well as the character embeddings work well when concatenated with the word embeddings. This configuration is what is kept in our final model. We further continue our experiments by turning on / off different modules such as word embedding, sense embedding, selective pickup from LSTM and bi-gram embedding. We found that the contribution of word embeddings, sense embeddings and selective pickup from LSTM are significant compared to the bigram module as shown by the weights at the 500th epoch in Table \ref{weights}. The bigram module works better whenever we do not consider the previously generated part-of-speech. So we kept the first three modules and discarded the bigram module from our final model. Our best model as shown in the last row of Table \ref{tricks} gives state of the art results. 

\section{Conclusion}
In this paper, we propose an improved neural sequence labelling architecture by leveraging from additional linguistic information such as polysemy, bigrams, character level knowledge and morphological features. Benefitting from such adequately captured linguistic information, we can assemble a considerably more compact model, hence yielding much better training time without loss of effectiveness. To avoid feature collision we performed an extensive ablation study where we produced an optimal model structure along with an optimal set of features. Our best model achieved state of the art results on the POS tagging, NER and chunking benchmark datasets and at the same time remains four times faster to train than the best performing model currently available. Our experimental results show that multiple linguistic features and their proper inclusion significantly boosted our model performance.

\bibliographystyle{IEEEtran}
\bibliography{reference}
\end{document}